\documentclass{article}

 \usepackage[final]{nips_2018}

\usepackage[utf8]{inputenc} 
\usepackage[T1]{fontenc}    
\usepackage{hyperref}       
\usepackage{url}            
\usepackage{booktabs}       
\usepackage{amsfonts}       
\usepackage{amsmath}
\usepackage{nicefrac}       
\usepackage{microtype}      
\usepackage{graphicx}
\usepackage{adjustbox}
\usepackage{natbib}
\setcitestyle{square,numbers}

\usepackage{subcaption}

\hypersetup{
    colorlinks = true,
    citecolor= blue
}

\title{Deep Ensemble Tensor Factorization for Longitudinal Patient Trajectories Classification}

\author{
  Edward~De Brouwer\thanks{Corresponding author.} \\
  ESAT-STADIUS\\
  KU LEUVEN\\
  Leuven, 3001, Belgium \\
  \texttt{edward.debrouwer@esat.kuleuven.be} \\
  \And
  Jaak Simm \\
  ESAT-STADIUS \\
  KU LEUVEN \\
  Leuven, 3001, Belgium  \\
  \texttt{jaak.simm@esat.kuleuven.be}\\
  \AND
  Adam Arany \\
  ESAT-STADIUS \\
  KU LEUVEN \\
  Leuven, 3001, Belgium\\
   \texttt{adam.arany@esat.kuleuven.be}\\
  \And
  Yves Moreau \\
   ESAT-STADIUS \\
   KU LEUVEN \\
  Leuven, 3001, Belgium\\
  \texttt{moreau@esat.kuleuven.be}\\
}

\begin{document}

\maketitle

\begin{abstract}
	We present a generative approach to classify scarcely observed longitudinal patient trajectories. The available time series are represented as tensors and factorized using generative deep recurrent neural networks. The learned factors represent the patient data in a compact way and can then be used in a downstream classification task. For more robustness and accuracy in the predictions, we used an ensemble of those deep generative models to mimic Bayesian posterior sampling. We illustrate the performance of our architecture on an intensive-care case study of in-hospital mortality prediction with 96 longitudinal measurement types measured across the first 48-hour from admission. Our combination of generative and ensemble strategies achieves an AUC of over 0.85, and outperforms the SAPS-II mortality score and GRU baselines.
\end{abstract}

\section{Introduction}

Envisioned as key tool for precision medicine, the computational analysis of patient trajectories has recently been a major focus of interest \citep{Jensen2014,Soleimani,Razavian2016}. In particular, patient trajectories are expected to be of great importance in diseases that evolve over long time periods (\emph{e.g.}, chronic diseases such as diabetes or multiple sclerosis) or with highly patient-specific medical trajectory patterns (\emph{e.g.}, Intensive Care Unit (ICU) data). However, despite the increasing literature on medical health records mining \citep{Shickel}, there is still a lack of methods for modeling this particularly challenging type of data in a natural way.

Patient trajectories usually consist of complex sets of longitudinal measurements (\emph{e.g.}, blood glucose), medical events (\emph{e.g.}, onset of comorbidities), or patient covariates (\emph{e.g.}, gender). They are relevant for a broad range of medical tasks, such as (1) patient segmentation \citep{Zhu2016,Nigam}, (2) prognosis \citep{Yang}, or (3) treatment optimization \citep{Raghu}. However, their statistical analysis is challenging because of the inherent properties of the data \citep{Lee2014,Wells2013}. First, the temporal series are often scarcely observed (\emph{i.e.}, only a few percent of possible instances are actually measured) and irregularly sampled (\emph{e.g.}, a multiple sclerosis patient typically has a medical visit every 6 months). Second, the observation pattern is informative in itself as it reflects, among other things, the need for additional or fewer medical visits, and thus the state of the patient. Lastly, observations are noisy and because of the complexity of the diseases, there is seldom a natural way to align patients on a common time scale. Those limitations typically prevent us from directly feeding this type of data into classical supervised learning methods. 

In this work, we propose a modeling strategy that addresses the aforementioned issues in a direct way. We consider that the observed measurements are generated by a low-dimensional hidden temporal process that summarizes the health state of the patient at each point in time. The noisy observations are then interpreted as outputs, which results in a natural way of handling missing values. For modeling complex temporal dependencies in the data, and to address the trajectories alignment issue, we rely on a recurrent neural network architecture to generate the observations. We further boost the performance of our model by using an ensemble method to approximate the Bayesian sampling of the posterior of the predictions.



\section{Previous work}


The machine learning community has recently started to address the challenging problem of patient trajectory modelling \citep{Rajkomar2018, Shickel,Ranganath,Choia,Ghassemi2015,Pham2016,Yoon2016}. \citet{Choi} proposed Doctor AI, a GRU-based architecture to predict medical event at the next visit. In the specific task of patient trajectory classification, \citet{Lipton} proposed a LSTM-based model fed by imputed data concatenated with an observation mask. The closest work to our approach is the one of \citet{Che2018} who designed an extended GRU-cell for dynamic imputation mechanism that is trained for classification only. In contrast, we propose a model that, in addition to classification, generates the observed trajectories. Others approaches also include convolutional networks \citep{Cui}. Several recent works have also focused on the specific task of in-hospital mortality prediction in intensive care \citep{Che2018,Awad2017}.

\section{Methods}
\subsection{Data representation: \emph{Tensorization}}

The data we aim at analyzing typically consists of multiple longitudinal measurements for each patient together with their time labels. We first discretize the time into bins with high granularity resulting in minimal loss of information. Measurements falling in the same bin are either averaged or summed depending on the specific measurement type. 

We then represent the $M$ types of temporal medical measurements of $N$ patients over $T$ time steps as an order-3 tensor $\mathcal{Y}$ of dimension $N \times M \times T$. 
In the applications we focus on, this tensor is usually scarcely observed, resulting in a low fill rate of just a few percents.

On top of longitudinal measurements, static information about patients is also available. We write the matrix of static patient covariates as $\mathbf{X}$ of dimensions $N \times K$ with $K$ the number of covariates. This matrix is assumed to be fully observed. That is, all covariates are known for each patient.

Finally, each patient is assigned a label that defines its class. We write $\mathbf{z} \in \mathbb{Z}^N$ the vector of class labels of all patients.

\subsection{Model definition}
The learning objective is to correctly classify the patient labels $\mathbf{z}$ using temporal information $\mathcal{Y}$ and static information $\mathbf{X}$. For this purpose, we proceed in two joint tasks: deep factorization of the tensor and classification of the patients based on the retrieved latent factors.

Regarding the factorization task, we assume that the temporal (tensor) observations of each patient $i$ are generated by a $D$-dimensional latent process $\mathbf{h}^i[t] \in \mathbb{R}^D$ and the corresponding measurement functions $g^j(\cdot)$:

\begin{equation}
 \mathcal{Y}_{i,j,t} = g^j (\mathbf{h}^i[t]) + \epsilon  \quad \quad \text{with}\quad  \quad \epsilon \sim \mathcal{N}(0,\sigma^2)
 \label{eq:samples}
 \end{equation}

This individual latent process can be interpreted as the hidden health status of the patient, conditioned on which the observations are generated. The model decomposes the tensor into temporal patient specific factors $h^i[t]$ and measurements specific functions $g^j(\cdot)$. We assume the first latent factors $h^i[0]$ are generated from the static covariates $\mathbf{X}_i \in \mathbb{R}^F$ (with generating function $\beta$ and noise $\epsilon$) and that these factors then transition over time according to an unknown process $v$ and noise $\xi$:

\begin{equation}
\mathbf{h}^i[0] \sim \beta(\mathbf{X}_i)+ \epsilon \quad \text{,} \quad
\mathbf{h}^i[t] = v(\mathbf{h}^i[t-1]) + \xi \quad \text{with} \quad
\epsilon \sim \mathcal{N}(0, \mathbf{\Sigma}_{\epsilon}), \quad
\xi \sim \mathcal{N}(0, \mathbf{\Sigma}_{\xi})
\label{eq:latents}
\end{equation}

We further assume that the binary patients labels are also generated by their last hidden health process value $\mathbf{h}^i[T]$ through a mapping $w$. Specifically, $p(z_i=1)=w(\mathbf{h}^i[T]) $.

\subsection{Inference and learning}

As we want to account for complex generating functions, exact inference of the latent temporal process is intractable. We therefore adopt a recurrent neural network approach that computes an approximate inference of the patient hidden process.


The first latents $h^i[0]$ are generated from a nonlinear mapping of the static covariates as shown in \eqref{eq:latents}. We then feed a GRU-based network with the longitudinal observations concatenated with observations mask $\mathcal{M}$. When the samples are not observed, we impute the missing observations with the predictions of the previous time steps.

Specifically, at each time step $t$, for patient $i$, we feed the GRU cell with a vector $\mathbf{y}^*$ such that

$$\mathbf{y}^*=[\mathbf{y} ;  \mathcal{M}_{i,:,,t}] \quad \text{with} \quad y_j=\mathcal{Y}_{i,j,t} \quad \text{if} \quad  \mathcal{M}_{i,j,t}=1 \quad \text{and}  \quad y_j=\hat{y}_j \quad \text{otherwise}$$

where $\mathcal{M}_{i,j,t}=1$ if sample $\mathcal{Y}_{i,j,t}$ is observed and $0$ otherwise. The network is then fed with the observation pattern and we let the GRU design its own imputation strategy when the sample is not available. Note that we design the network such that it generates the observations based on the latents at each time step as in \eqref{eq:samples}. We then jointly train a classifier $w$ on the last hidden vector of each patient $h^i[T]$ for the labels.

\begin{figure}[htbp]
  \centering
    \includegraphics[width=12cm]{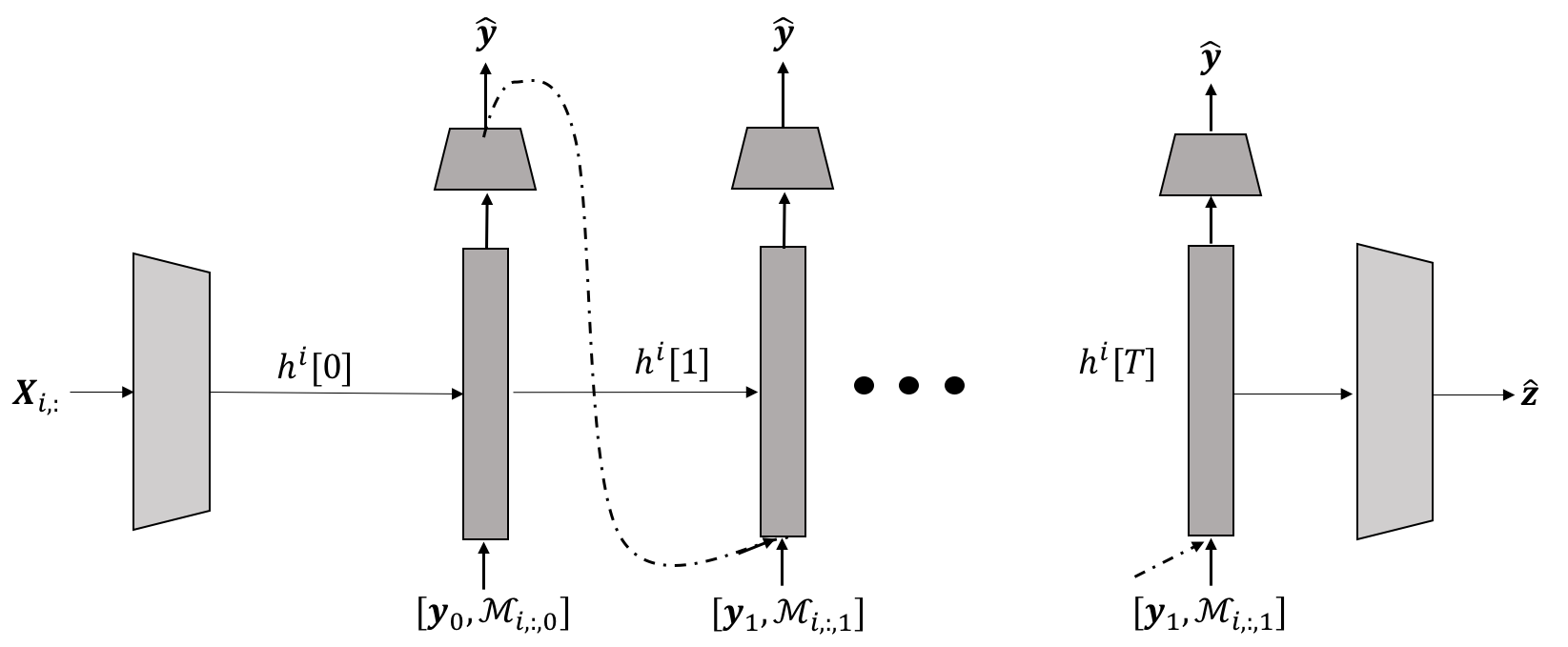} 
    \caption{Model architecture. Rectangles represent GRU cells unfolded over time. The network both generates class labels and the observed temporal trajectories.}
    \label{fig:GRU}
\end{figure}

Two objectives coexist in this model architecture: the ability to reconstruct the observed patient trajectories based on the hidden process and the classification performance. We train our model using the mixing of those two goals with a hyperparameter $\gamma$. The overall loss to optimize is then

$$ \mathcal{L}oss = \gamma  \lVert \mathcal{Y}-\mathcal{\hat{Y}}_{\mathbf{w}} \rVert^2+ (1-\gamma) H(\mathbf{z},\mathbf{\hat{z}}_{\mathbf{w}} ) + \lambda \lVert \mathbf{w} \rVert^2,$$
where $H(\cdot,\cdot)$ stands for the cross entropy loss and $\mathbf{w}$ are the weights of the model. A visual representation of the network architecture is presented in Figure~\ref{fig:GRU}. The generative approach mainly presents two advantages: the latent vectors are imposed to be representative of the whole observed trajectory and they provide a more natural way to deal with missing input samples. 

For better predictive accuracy, we use a performance driven ensemble learning method to approximate Bayesian posterior sampling of the predictions. Practically, we train a large number of models with hyperparameters sampled from some prior distribution. We then select the best models according to validation performance and average their predictions.

\section{Experiments}
\subsection{Case study definition}
We apply our methodology to an intensive care unit case study. We used the publicly available MIMIC III data set that contains longitudinal measurements for more than 40,000 critical care patients \citep{Johnson2016a}. The objective of our case study is to predict, based on longitudinal data of a patient in a 48 hours observation window, in-hospital mortality for that individual patient. We selected a subset of 20,000 patients with at least 48 hours of hospital stay. For each patient, we selected 96 different longitudinal measurements types divided in 4 main categories: lab measurements, inputs to patients, outputs collected from patients, and drug prescriptions. The full list is available in the supplementary material. The selected time series were scarcely observed leading to a filling rate of 5.9\% for tensor $\mathcal{Y}$. We also selected static patient covariates ($\mathbf{X}$), such as age, admission type, and main ICD10 diagnose for the admission.

\subsection{Our models and baselines}

We consider the following baselines: the SAPS-II\footnote{We restricted the SAPS-II severity score to the variables that were available in the data subset under consideration.} severity score and a missingness-informed GRU baseline with smart imputation. We call this architecture \emph{GRU-imputed}. 
 In contrast to our proposed method, this architecture is trained for classifying the in-hospital mortality labels only. It therefore does not learn any hidden process generating the required observations. At each time step \emph{GRU-imputed} is fed with a vector of observations (with missing values imputed to their means as suggested by \citet{Lipton} and \citet{Che2018}.) concatenated with an observation mask and the elapsed time since last observed sample. The SAPS-II is a static severity score widely used in clinical practice. 

We then trained 200 of our models with hyperparameters sampled from the following priors: $ \gamma \sim \mathcal{U}nif(0,0.1)$ and $\log(\lambda) \sim \mathcal{U}nif(-8,-2)$ on the training set. We then ranked the models based on their performance on the validation set, selected an ensemble of the best 20, averaged their predictions and report the performance on an held out test set. The same data splitting was used to tune and evaluate all models.

\subsection{Performance}

\begin{figure}[htbp]
\centering
\begin{subfigure}{.5\textwidth}
  \centering
 \includegraphics[width=7.5cm]{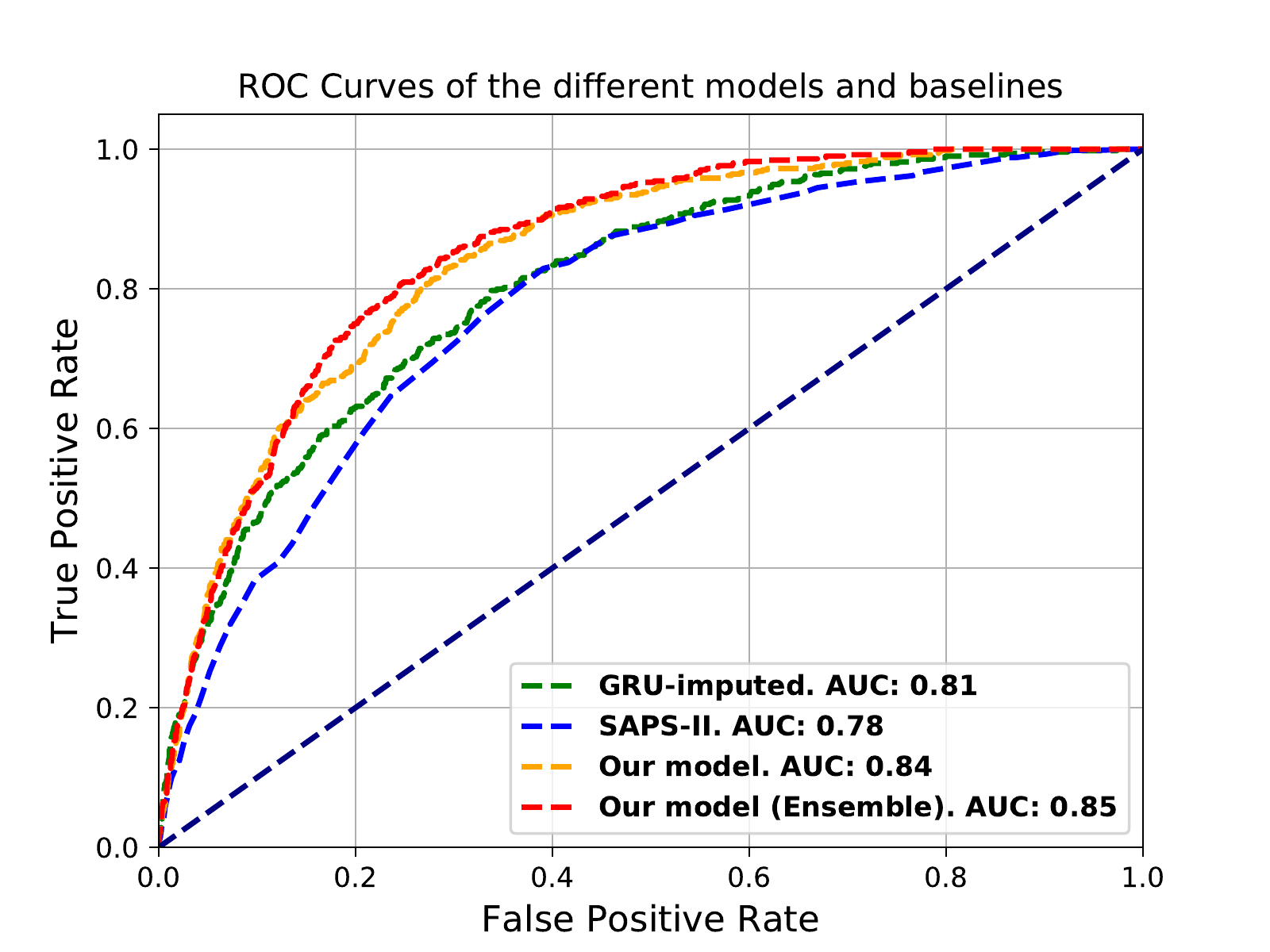} 
 \caption{ROC curves and AUC for the different models}
\label{fig:sub1}
\end{subfigure}%
\begin{subfigure}{.5\textwidth}
  \centering
 \includegraphics[width=7.5cm]{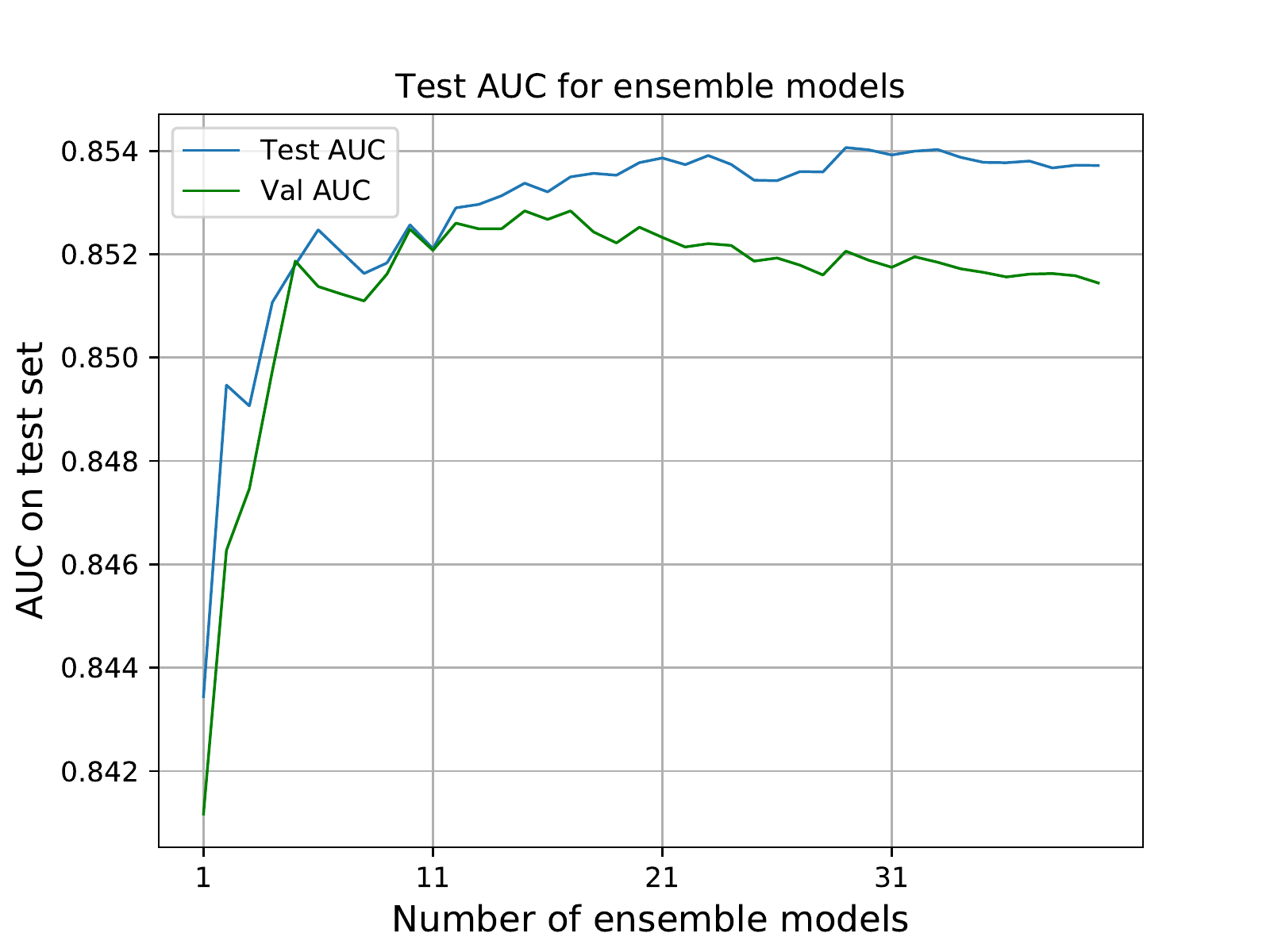} 
  \caption{Impact of the ensemble strategy on the performance}
  \label{fig:sub2}
\end{subfigure}
\caption{Results of the models considered}
\label{fig:results}
\end{figure}

The performance of both baselines and our method is presented on Figure~\ref{fig:sub1}. Our methodology outperforms the proposed baselines. Furthermore, we notice an increase from 0.842 to 0.855 in AUC due to the ensemble strategy. Impact of the number of models in the ensemble is presented in Figure~\ref{fig:sub2}. We observe that few models are required to obtain significant performance improvement.

\section{Conclusion}
We propose to consider multivariate longitudinal patient trajectories as a higher-order tensor that is factorized using deep recurrent neural networks. The temporal factors of each patient are then used for in-hospital mortality prediction. The performance of the proposed architecture shows that the generative approach outperforms both classification-only and static models. 

\vfill

\bibliographystyle{plainnat}
\bibliography{library.bib}

\section{Supplementary material}

\subsection{Retained features for the intensive care case study}

In table \ref{tab:features}, we present the longitudinal features retained for the training of our models.

\begin{table}[htbp]
    \begin{adjustbox}{width=\textwidth}
   \begin{tabular}{| l | c | c | r | } 
     \hline
     \multicolumn{4}{|c|}{\textbf{Retained Features}} \\
     \hline
      Lab measurements   & Inputs & Outputs & Prescriptions\\
      \hline
      Anion Gap & Potassium Chloride & Stool Out Stool & D5W \\
      Bicarbonate & Calcium Gluconate & Urine Out Incontinent & Docusate Sodium \\
      Calcium, Total & Insulin - Regular & Ultrafiltrate Ultrafiltrate & Magnesium Sulfate \\
      Chloride & Heparin Sodium & Gastric Gastric Tube & Potassium Chloride \\
       Glucose & K Phos & Foley & Bisacodyl \\
       Magnesium & Sterile Water & Void & Humulin-R Insulin \\
       Phosphate & Gastric Meds & TF Residual & Aspirin \\
       Potassium & GT Flush &Pre-Admission & Sodium Chloride 0.9\%  Flush \\
       Sodium & LR & Chest Tube 1 & Metoprolol Tartrate \\
       Alkaline Phosphatase & Furosemide (Lasix) & OR EBL & \\
       Asparate Aminotransferase & Solution & Chest Tube 2 & \\
        Bilirubin, Total & Hydralazine & Fecal Bag & \\
       Urea Nitrogen & Midazolam (Versed) & Jackson Pratt 1 & \\
       Basophils & Lorazepam (Ativan) & Condom Cath & \\
       Eosinophils & PO Intake & & \\
       Hematocrit & Insulin - Humalog & & \\
       Hemoglobin & OR Crystalloid Intake & & \\
       Lymphocytes & Morphine Sulfate & & \\
       MCH & D5 1/2NS & & \\
       MCHC & Insulin - Glargine & & \\
       MCV & Metoprolol & & \\
       Monocytes & OR Cell Saver Intake & & \\
       Neutrophils & Dextrose 5\% & & \\
       Platelet Count & Norepinephrine & & \\
       RDW & Piggyback & & \\
       Red Blood Cells & Packed Red Blood Cells & & \\
       White Blood Cells & Phenylephrine & & \\
       PTT & Albumin 5\% & & \\
       Base Excess & Nitroglycerin & & \\
       Calculated Total CO2 & KCL (Bolus) & & \\
       Lactate & Magnesium Sulfate (Bolus) & & \\
       pCO2 & & & \\
       pH & & & \\
       pO2 & & & \\
       PT & & & \\
       Alanine Aminotransferase & & & \\
       Albumin & & & \\
       Specific Gravity & & & \\
      \bottomrule
   \end{tabular}
   \end{adjustbox}
    \caption{Retained longitudinal features in the intensive care case study.}
   \label{tab:features}
\end{table}

\end{document}